\documentclass{article}

\usepackage{arxiv}

\usepackage[utf8]{inputenc} 
\usepackage[T1]{fontenc}    
\usepackage{hyperref}       
\usepackage{url}            
\usepackage{booktabs}       
\usepackage{amsfonts}       
\usepackage{nicefrac}       
\usepackage{microtype}      
\usepackage{lipsum}
\usepackage{graphicx}
\graphicspath{ {./images/} }

\title{Delayed Feedback in Kernel Bandits}

\author{
 Sattar Vakili \\
  MediaTek Research \\
  \texttt{sattar.vakili@mtkresearch.com} \\
   \And
 Danyal Ahmed\\
  MediaTek Research \\
\texttt{danyal.ahmed@mtkresearch.com} \\
  \And
 Alberto Bernacchia \\
  MediaTek Research \\
  \texttt{alberto.bernacchi@mtkresearch.com} \\
  \And
 Ciara Pike-Burke \\
  Imperial College London \\
\texttt{c.pike-burke@imperial.ac.uk } \\
}

\usepackage{amsmath}
\usepackage{amssymb}
\usepackage{mathtools}
\usepackage{amsthm}

\usepackage{algorithm}
\usepackage{algpseudocode}

\def\Xx{\mathbb{X}}
\def\Rr{\mathbb{R}}
\def\Nn{\mathbb{N}}
\def\E{\mathbb{E}}

\def\Xc{\mathcal{X}}
\def\Oc{\mathcal{O}}
\def\Rc{\mathcal{R}}
\def\Ec{\mathcal{E}}
\def\Ic{\mathcal{I}}
\def\Hc{\mathcal{H}}

\def\Xb{\mathbf{X}}
\def\Kb{\mathbf{K}}
\def\yb{\mathbf{y}}
\def\Ib{\mathbf{I}}
\def\kb{\mathbf{k}}

\def\sigmat{\tilde{\sigma}}
\def\mut{\tilde{\mu}}
\def\dt{\tilde{d}}
\def\Oct{\tilde{{\mathcal{O}}}}
\def\Xbt{\tilde{\Xb}}
\def\ybt{\tilde{\yb}}
\def\betat{\tilde{\beta}}

\def\nn{\nonumber}

\usepackage[round]{natbib}
\usepackage{color}
\usepackage{subfig}
\hypersetup{
    colorlinks=true,
    linkcolor=magenta,
    citecolor =cyan,
    filecolor=magenta,      
    urlcolor=magenta,
}

\theoremstyle{plain}
\newtheorem{theorem}{Theorem}[section]

\newtheorem{lemma}[theorem]{Lemma}
\newtheorem{corollary}[theorem]{Corollary}
\theoremstyle{definition}

\newtheorem{assumption}[theorem]{Assumption}
\theoremstyle{remark}

\begin{document}
\maketitle

\begin{abstract}
    Black box optimisation of an unknown function from expensive and noisy evaluations is a ubiquitous problem in machine learning, academic research and industrial production. An abstraction of the problem can be formulated as a kernel based bandit problem (also known as Bayesian optimisation), where 
    a learner aims at optimising a kernelized function through sequential noisy observations. The existing work predominantly assumes feedback is immediately available; an assumption which fails in many real world situations, including recommendation systems, clinical trials and hyperparameter tuning.
    We consider a kernel bandit problem under stochastically delayed feedback, and propose an algorithm with $\Oct(\sqrt{\Gamma_k(T) T}+\E[\tau])$ regret, where $T$ is the number of time steps, $\Gamma_k(T)$ is the maximum information gain of the kernel with $T$ observations, and $\tau$ is the delay random variable. This represents a significant improvement over the state of the art regret bound of $\Oct(\Gamma_k(T)\sqrt{ T}+\E[\tau]\Gamma_k(T))$ reported in~\citet{verma2022bayesian}. In particular, for very non-smooth kernels, the information gain grows almost linearly in time, trivializing the existing results. 
    We also validate our theoretical results with simulations. 
\end{abstract}


\section{Introduction}
The kernel bandit problem is a flexible framework which captures the problem of learning to optimise an unknown function through successive input queries. 
Typically, the game proceeds in rounds where in each round the learner selects an input point to query and then immediately receives a noisy observation of the function at that point. This observation can be used immediately to improve the learners decision of which point to query next. 
Due to its generality, the kernel bandit problem has become very popular in practice. In particular, it enables us to sequentially learn to optimise a variety of different functions without needing to know many details about the functional form.

However, in many settings where we may want to use kernel bandits, we also have to deal with delayed observations. For example, consider using kernel bandits to sequentially learn to select the optimal conditions for a chemical experiment. The chemical reactions may not be instantaneous, but instead take place at some random time in the future. If we start running a sequence of experiments, we can start new experiments before receiving the stochastically delayed feedback from the previous ones. However, in this situation we have to update the conditions for future experiments before receiving all the feedback from previous experiments. Similar situations arise in recommendation systems, clinical trials and hyperparameter tuning, so it is of practical relevance that we design kernel bandit algorithms that are able to deal with delayed feedback.
Moreover, a big challenge for existing kernel bandit algorithms is the computational complexity. Generally speaking, in each round $t$, algorithms for kernel bandits require fitting a kernel model to the $t$ observed data points which can have an $\Oc(t^3)$ complexity. To reduce the complexity, there has been a recent interest in considering batch versions of kernel bandit algorithms. These algorithms select $\tau$ input values using the same model then update the model after receiving all observations in the batch. This corresponds to a delay of at most $\tau$ in receiving each observation, and thus can be thought of as an instance of delayed feedback in kernel bandits.

In this paper, we study the kernel bandit problem with stochastically delayed feedback. 
We propose Batch Pure Exploration with Delays (BPE-Delay) ---an adaptation of the Batch Pure Exploration algorithm \citep{li2022gaussian} to the delayed feedback setting---, and show that, under mild assumptions on the unknown delay distribution, BPE-Delay achieves near optimal regret. Indeed, we prove that BPE-Delay enjoys regret scaling as $\Oct(\sqrt{\Gamma_k(T) T}+\E[\tau])$,\footnote{We use the notation $\Oc$ and $\Oct$ to denote order and order up to logarithmic factors, respectively.} where $T$ is the number of time steps, $\Gamma_k(T)$ is the maximum information gain of the kernel $k$ with $T$ observations (see Section~\ref{sec:Related}), and $\tau$ is the delay random variable. This result essentially shows that the impact of delayed feedback on the regret of this algorithm is independent of the horizon $T$ and the problem parameter $\Gamma_k(T)$. This desirable property means that as $T$ or $\Gamma_k(T)$ increase, the impact of the delay remains the same.

We note that for linear models, ${\Gamma_k(T)}=\Oc(d\log(T))$, where $d$ is the input dimension. In the case of very smooth kernels such as Squared Exponential (SE), ${\Gamma_k(T)}=\text{poly}\log(T)$. However ${\Gamma_k(T)}$ can become arbitrary close to linear in $T$ for less smooth kernels. For example, in the case of a Mat{\'e}rn kernel with smoothness parameter $\nu$, we have $\Gamma_k(T)=\Oct(T^{\frac{d}{2\nu+d}})$~\citep{vakili2020information}. 
Therefore, our results represent a significant theoretical improvement over the existing work on kernel bandits with delayed feedback, where an $\Oct(\Gamma_k(T)\sqrt{ T}+\E[\tau]\Gamma_k(T))$ regret bound was shown for an algorithm based on upper confidence bound (UCB) acquisition of observation points~\citep{verma2022bayesian}.
In particular, for non-smooth kernels, our result reduces the delay penalty from possibly near $\Oct(\E[\tau]T)$ to just $\Oct(\E[\tau])$.
We demonstrate in Section~\ref{sec:exp} that these theoretical improvements are also visible experimentally. 
In addition, when applied to the special case of linear bandits (a kernel bandit problem with linear kernel), our results improve the dependency of the delay related term in the regret bound on the input dimension by a $d^{3/2}$ factor compared to the state of the art in \citet{howson2022delayed}. A detailed comparison with related work is provided in Section~\ref{sec:Related}.

\section{Background and Related Work}\label{sec:Related}

In this section, we give an overview of the background on kernel bandits with immediate feedback and delayed feedback in simpler stochastic bandit formulations (namely in $K$-armed and linear bandits). We then provide a detailed comparison with the most related work on kernel bandits with delayed feedback.  

\paragraph{Kernel bandits with immediate feedback.} 
In the typical kernel bandit setting with immediate feedback, 
classic algorithms such as selecting the query points based on upper confidence bounds (GP-UCB)~\citep{srinivas2010gaussian} and Thompson Sampling~\citep{Chowdhury2017bandit} achieve a regret bound of $\Oct({\Gamma_k(T)}\sqrt{T})$. This regret bound is suboptimal~\citep[see,][for details]{vakili2021open}, and can be improved to $\Oct(\sqrt{{\Gamma_k(T)}T})$ using more recent work. In particular, Batch Pure Exploration (BPE) introduced in~\citet{li2022gaussian} and Threshold Domain Shrinking (GP-ThreDS) proposed in~\citet{salgia2021domain} achieve this improved regret bound. Here, ${\Gamma_k(T)}$ is a kernel specific complexity term, which can be interpreted as the \emph{information gain} or the \emph{effective dimension}.

\paragraph{Information gain and effective dimension.} While the feature space representation of common kernels is infinite dimensional, with a finite data set, only a finite number of features have a significant effect on the kernel based regression. That leads to the definition of the effective dimension. In particular, consider a finite set $\Xb_T=\{X_1,\dots, X_T\}$ of observation points and a positive definite kernel $k:\Xc\times\Xc\rightarrow \Rr$. Let $\Kb_{\Xb_T}=[k(X_t,X_{t'})]_{t,t'=1}^T$ be the kernel matrix resulting from the pairwise kernel values between the observation points. The effective dimension for a given kernel and observation set is often defined as~\citep{zhang2005learning, Valko2013kernelbandit} 
\begin{eqnarray}
\tilde{d}_k(T) = \mathrm{tr}\left(\Kb_{\Xb_T}(\Kb_{\Xb_T}+\lambda^2\Ib_T)^{-1}\right),
\end{eqnarray}
where $\lambda>0$ is a free parameter and $\Ib_T$ denotes an identity matrix of dimension $T$.

It is also useful to define a related notion of information gain. For this, assume $f$  is a centered Gaussian Process (GP) on the domain $\Xc$ with kernel $k$. Information gain then refers to the mutual information $\Ic(\yb_T; f)$ between the data values $\yb_T=[y_t=f(X_t)+\epsilon_t]_{t=1}^T$ and $f$ (assuming the surrogate GP distribution and a zero mean Gaussian noise $\epsilon_t$ with variance $\lambda^2$). From the closed form expression of mutual information between two multivariate Gaussian distributions~\citep{cover1999elementsold}, it follows that
$
\Ic(\yb_T; f)  = \frac{1}{2}\log\det\left(\Ib_T+\frac{1}{\lambda^2}\Kb_{\Xb_T} \right). \label{eq:mutual_info}
$
We then define the data-independent and kernel-specific \emph{maximum information gain} as follows \citep{srinivas2010gaussian}:
\begin{eqnarray}\label{def:gamma}
\Gamma_k(T) = \sup_{\Xb_T\subset\Xc}\Ic(\yb_T; f). \label{eq:gamma}
\end{eqnarray}
It is known that the information gain and the effective dimension are the same up to logarithmic factors. Specifically,
we have $\dt_k(T)\le \Ic(\yb_T;f)$, and $\Ic(\yb_T; f)=\Oc(\dt_k(T)\log(T))$~\citep{Calandriello2019Adaptive}.
For specific kernels, explicit bounds on $\Gamma_{k}(T)$ are derived in~\citet{srinivas2010gaussian, vakili2021uniform, vakili2020information}. We report our regret bounds in terms of information gain that can be easily replaced by the effective dimension, up to logarithmic factors. 

\paragraph{Regret lower bounds.} For commonly used SE and Mat{\'e}rn kernels, \citet{Scarlett2017Lower} derived lower bounds on regret (in the immediate feedback setting). In particular, they showed $\Omega(\sqrt{T(\log(T))^{d/2}})$ and $\Omega(T^{\frac{\nu+d}{2\nu+d}})$ lower bounds on regret, in the case of SE and Mat{\'e}rn kernels, respectively. These bounds are matched by the regret bounds for GP-ThreDS~\citep{salgia2021domain} and BPE~\citep{li2022gaussian}, up to logarithmic factors.

\paragraph{Delayed feedback in stochastic bandits.}
Delayed feedback has been studied significantly in the stochastic bandit problem where there are $K$ independent arms. 
\citet{joulani2013online} provided a queue based wrapper algorithm which, when applied with any bandit algorithm, leads to an additive penalty of $\E[\tau]$ in the regret compared to the non-delayed setting. They also showed that using a UCB algorithm \citep{auer2002finite} just with the observed data would lead to a similar regret bound of $\Oct(\sqrt{KT} + \E[\tau])$. \citet{mandel2015queue} also provided a queue based algorithm with the same regret bound. 
\citet{vernade2017stochastic} developed a UCB algorithm to deal delayed feedback where some observations are censored if their delay exceeds a threshold.
We note that we cannot directly extend these queue based approaches to the kernel bandit problem where we have a large number of dependent arms. However we show that the same additive penalty can be maintained even in our more difficult setting. Additionally, while it is possible to extend the UCB based algorithms to our setting, in Section~\ref{sec:exp}, we show that our proposed algorithm performs better empirically than using a delayed version of GP-UCB.

There has also been work studying the impact of delayed feedback in generalised linear bandits. \citet{zhou2019learning} and \citet{howson2022delayed} provided adaptations of optimistic (UCB) algorithms to account for delayed feedback with sub-exponential delays. \citet{zhou2019learning} 
obtained a regret bound scaling with  $\Oct(d\sqrt{T}+\sqrt{dT\E[\tau]})$
, while \citet{howson2022delayed} 
obtained an improved regret bound of $\Oct(d\sqrt{T}+d^{\frac{3}{2}} \E[\tau])$, although here the delay penalty still depends on the dimension which is not the case for the delayed stochastic $K$-armed bandit problem. When applied to this setting, we show that our proposed algorithm removes this interaction between the dimension and the delay. Specifically, our results improve the delay related term in the regret bound with a $d^{\frac{3}{2}}$ factor in the special case of linear bandits (a kernel bandit problem with linear kernel). 
We also note that \citet{vernade2020linear} extended their work on delayed feedback with censoring to the contextual linear bandit setting, and \citet{dudik2011efficient} studied constant delays in the the contextual bandit setting, although these settings are not directly comparable to ours.

\paragraph{Delayed feedback in kernel bandits.}

The most relevant work to ours is \citet{verma2022bayesian} where the kernel bandit problem with stochastically delayed feedback was also considered. \citet{verma2022bayesian} proposed algorithms based on GP-UCB~\citep{srinivas2010gaussian} and Thompson sampling~\citep{Chowdhury2017bandit} in the delayed feedback setting. In these algorithms, referred to as GP-UCB-SDF and GP-TS-SDF (where SDF stands for Stochastic Delayed Feedback), the unavailable feedback due to delay are replaced by minimum function value (assuming it is known). 
They provided a regret bound for this algorithm of $\Oct(\Gamma_k(T)\sqrt{ T}+\E[\tau] \Gamma_k(T))$. This is an improvement over a naive application of the existing algorithms (which would lead to a regret bound of $\Oc({\Gamma_k({T}})\sqrt{T\E[\tau]})$), but still suffers from a scaling of the term involving the delay by $\Gamma_k(T)$. 
In comparison, our algorithm does not require the additional knowledge of the minimum function value (we simply disregard the unavailable observations). Furthermore, 
our results significantly improve upon \citet{verma2022bayesian}, by completely decoupling the impact of the delay and the problem  parameters. Our regret bounds are also order optimal and cannot be further improved for the cases where a lower bound on regret is known, in particular, the bounds given in~\citet{Scarlett2017Lower} for the SE and Mat{\'e}rn kernels (under the immediate feedback setting). 

Kernel bandits with batch observations can be considered as a special case of our delayed feedback framework, with constant (non-stochastic) delays. Specifically, the observations for the points in a batch are available with a fixed delay that is at most the length of the batch.  Notable examples are \citet{desautels2014parallelizing,vakili2021scalable, daxberger2017distributed, chowdhury2019batch} which are based on the \emph{hallucination} technique introduced in~\citet{desautels2014parallelizing}. In this technique, the unavailable observations are \emph{hallucinated} to be the kernel based prediction using only the available feedback. This is equivalent to keeping the prediction unaffected by the unavailable observations, while updating the uncertainty estimate using all selected observation points, including the ones with delayed observations~\citep{chowdhury2019batch}. In contrast, \citet{verma2022bayesian} set the unavailable observations to the minimum function value (assuming it is known). In our algorithm, we simply disregard the unavailable observations in the prediction, while using both variations of uncertainty estimate (one updated using all observation points, and one updated using only the observation points with available feedback). Details are given in Section~\ref{sec:algorithm}. Our regret bounds also offer a significant improvement over the batch setting, despite this being a significantly simpler setting due to the fixed and known delays. In particular, in the batch setting, the best known regret bounds by \citet{chowdhury2019batch} 
are $\Oct(\Gamma_k(T)\sqrt{T}+\tau\sqrt{T\Gamma_k(T)} )$, where with an abuse of notation to enable comparison with our results, we use $\tau$ for the batch size. Theorem~\ref{the:regret} improves upon both terms in this regret bound.


\section{Problem Definition}\label{sec:PF}

Consider a positive definite kernel $k:\Xc\times\Xc\rightarrow \Rr$ supported on a compact $d$-dimensional set $\Xc\subset \Rr^d$ . A Hilbert space $\Hc_k$ of functions on $\Xc$ equipped with an inner product $\langle\cdot, \cdot\rangle_{\Hc_k}$ is called a reproducing kernel Hilbert space (RKHS) with reproducing kernel $k$ if the following is satisfied. For all $x\in\Xc$, $k(\cdot, x)\in\Hc_k$, and for all $x\in\Xc$ and $f\in\Hc_k$, $\langle f,k(\cdot, x)\rangle_{\Hc_k}=f(x)$ (reproducing property).

We consider a kernel bandit problem. We assume there exists an unknown objective function $f:\Xc\rightarrow \Rr$ of interest in the RKHS of a known kernel $k$. This is a very general assumption, since the RKHS of common kernels can approximate almost all continuous functions on compact subsets of the Euclidean space~\citep{srinivas2010gaussian}. 
Our aim is to find an input $x^*$ which maximises the unknown function $f$, i.e. $x^*\in\arg\max_{x\in\Xc}f(x)$. In order to do this, we can sequentially query $f$ at a chosen sequence of observation points $X_t\in\Xc$, $t=1,2,\dots$, and receive the noisy feedback $y_t=f(X_t)+\epsilon_t$ where $\epsilon_t$ is a sequence of independently distributed sub-Gaussian noise with zero mean. 

We measure the performance of any procedure for this problem in terms of its \emph{regret}. The regret is defined as 
\begin{eqnarray}
\Rc(T)=\sum_{t=1}^Tf(x^*)-f(X_t)
\end{eqnarray}
and represents the cumulative amount that we loose by querying $f$ at $X_1,\dots, X_T$ rather than at an unknown maxima $x^*\in\arg\max_{x\in\Xc}f(x)$.

In order to make the problem tractable, we make the following assumptions which are common in the literature.
\begin{assumption}\label{ass:f_norm}
The RKHS norm of the objective function $f$ is bounded, i.e., $\|f\|_{\Hc_k}\le C_k$, for some $C_k>0$, where the notation $\|\cdot\|^2_{\Hc_k}=\langle\cdot,\cdot\rangle_{\Hc_k}$ is used for the RKHS norm.  
\end{assumption}

\begin{assumption}\label{ass:noise}
The observation noise terms $\epsilon_t$ are independent $\sigma$-sub-Gaussian random variables with zero mean.  That is, for all $t\in \Nn$, for all $\eta \in \Rr$, and for some $\sigma>0$, the moment generating function of $\epsilon_t$ satisfies $\E[\exp(\eta \epsilon_t)]\le \exp(\frac{\eta^2\sigma^2}{2})$.
\end{assumption}

\paragraph{Delayed feedback.} In this work, we are interested in the kernel bandit problem in a setting with stochastically delayed feedback. In this setting, at each time $t$, as well as generating the observation $y_t$, the environment also independently generates a stochastic delay $\tau_t\geq 0 $. The learner receives the feedback for the decision made at time $t$ at the observation time $t+\tau_t$. 
We assume that $\tau_1,\dots, \tau_T$ represent a sequence of independent and identically distributed sub-exponential random variables. For simplicity, we assume that the $\tau_t$ random variables are discrete, although it is easy to extend our results to continuous delays by considering the events that $\tau_t + t \leq s$ for all $s \geq t$. Our assumption on the delay distribution is formally stated below, and is the same as that commonly made in the literature \citep{zhou2019learning,howson2022delayed}.

\begin{assumption}\label{ass:sub-exp}
    The delays $\tau_t$ are i.i.d. sub-exponential random variables. That is, for all $t\in \Nn$, for some $\xi,b>0$, for all $|\eta|\le \frac{1}{b}$, the moment generating function of $\tau_t$ satisfies $\E\left[\exp\left(\eta (\tau_t-\E[\tau_t])\right)\right]\le \exp(\frac{\eta^2\xi^2}{2})$.
 \end{assumption}


\section{Confidence Bounds}

Kernel based regression provides powerful predictors and uncertainty estimates. Those could be used to form confidence intervals for the unknown objective function $f$, that are a crucial building block in developing algorithms for the kernel based bandit problem. 
Consider a set of observation points $\Xb_t=\{X_1,\dots,X_t\}$, and the corresponding vector of observation values $\yb_t=[y_1, \dots, y_t]^{\top}$. Recall $\Kb_{\Xb_t}=[k(X_s,X_{s'})]_{{s,s'=1}}^t$ the positive definite kernel matrix resulting from the pairwise kernel values between the observation points. We have the following predictor and uncertainty estimate for $f$, which can be interpreted as the mean and variance of a surrogate GP model for $f$ with kernel $k$, respectively \citep[e.g., see,][]{Rasmussen2006, Kanagawa2018} 
\begin{eqnarray}\nn\label{GP_reg}
\mu_{\Xb_t,\yb_t}(x) &=& \kb^{\top}_{\Xb_t}(x)(\Kb_{\Xb_t}+\lambda^2\Ib_t)^{-1}\yb_t\\\nn
\sigma^2_{\Xb_t}(x) &=& k(x,x)-\kb^{\top}_{\Xb_t}(x)(\Kb_{\Xb_t}+\lambda^2\Ib_t)^{-1}\kb_{\Xb_t}(x),~~~
\end{eqnarray}
where $\kb_{\Xb_t}(x)=[k(x_1,x), \dots,k(x_t,x)]^{\top}$ is the kernel values vector between observation points and the point of interest and $\lambda>0$ is a free parameter.

Equipped with the above predictor and uncertainty estimate, we have the following confidence bounds. 

\begin{lemma}\label{lem:conf-single}[\citep{vakili2021optimal}]
Consider the predictor and uncertainty estimate given in~\eqref{GP_reg}. Assume the observation set $\Xb_t$ is selected independent of the observation noise $\epsilon_s$, $s=1,\dots, t$. Under Assumptions~\ref{ass:f_norm} and~\ref{ass:noise}, the following inequalities each hold with probability $1-\delta$ for any fixed $x\in\Xc$,
\begin{eqnarray}\nn
f(x) \le \mu_{\Xb_t,\yb_t}(x)+\beta(\delta)\sigma_{\Xb_t}(x),\\\nn
\medskip
f(x)  \ge \mu_{\Xb_t,\yb_t}(x)-\beta(\delta)\sigma_{\Xb_t}(x),
\end{eqnarray}
where $\beta(\delta) = C_k+\frac{\sigma}{\lambda}\sqrt{2\log(\frac{1}{\delta})}$, $C_k$ and $\sigma$ are the parameters given in Assumptions~\ref{ass:f_norm} and~\ref{ass:noise}, and $\lambda$ is the parameter in~\eqref{GP_reg}.
\end{lemma}

If the domain $\Xc$ is finite, then uniform confidence bounds
readily follow from this result via a union bound, and $\frac{\delta}{|\Xc|}$
can be substituted for $\delta$. For continuous domains, a technical discretization argument can be used to extend Lemma~\ref{lem:conf-single} to a uniform bound under the following continuity assumption.

\begin{assumption}\label{ass:disc}
For each $t\in\Nn$, there exists a discretization $\Xx$ of $\Xc$ such that, for any $f\in \Hc_k$ with $\|f\|_{\Hc_k}\le C_k$, we have $f(x) - f([x])\le \frac{1}{t}$, where $[x] = {\arg}{\min}_{ x'\in \Xx}||x'-x||_{l^2}$ is the closest point in $\Xx$ to $x$, and $|\Xx|\le cC_k^dt^{d}$, where $c$ is a constant independent of $t$ and $C_k$.
\end{assumption}
Assumption~\ref{ass:disc} is a mild and technical assumption that holds for typical kernels such as SE and Mat{\'e}rn~with $\nu>1$~\citep{srinivas2010gaussian, Chowdhury2017bandit, vakili2021optimal}.

\begin{lemma}[A special case of Corollary $3.7$ in~\citet{vakili2022improved}]\label{lem:conf-unif}
    Under the setting and assumptions of Lemma~\ref{lem:conf-single}, under Assumption~\ref{ass:disc}, the following inequalities each hold uniformly in $x\in\Xc$ with probability at least $1-\delta$,
    \begin{eqnarray}\nn
    f(x) \le \mu_{\Xb_t,\yb_t}(x)+\frac{2}{t}+\beta'_{\delta}(t)(\sigma_{\Xb_t}(x)+\frac{2}{\sqrt{t}}),\\\nn
    f(x) \ge \mu_{\Xb_t,\yb_t}(x)-\frac{2}{t}-\beta'_{\delta}(t)(\sigma_{\Xb_t}(x)+\frac{2}{\sqrt{t}}), 
\end{eqnarray}
where $\beta'_{\delta}(t) = \beta(\frac{\delta}{2\Gamma_t})$, $\Gamma_t=c(\tilde{C}_k(\frac{\delta}{2}))^dt^d$, $\tilde{C}_k(\delta) = C_k+\frac{\max\{\sigma,k_{\max}\}\sqrt{t}}{\lambda}\sqrt{2\log(\frac{2t}{\delta})}$, $k_{\max}=\max_{x\in\Xc}k(x,x)$. 
\end{lemma}

Lemma~\ref{lem:conf-unif} uses a high probability bound $\tilde{C}_k(\delta)$ on the RKHS norm of $\mu_{\Xb_t,\yb_t}$ (that is a random variable due to the random noise), together with applying the continuity assumptions to $\mu_{\Xb_t,\yb_t}$ and $\sigma_{\Xb_t}$ to construct the confidence bounds. Effectively, the multiplier of the width of the confidence intervals, $\beta'_{\delta}(t)$, scales as 
\begin{eqnarray}
    \beta'_{\delta}(t)=\Oc\left(C_k+\frac{\sigma}{\lambda}\sqrt{d\log(\frac{t C_k}{\delta})}\right).
\end{eqnarray}

In the kernel bandit with delayed feedback setting, some observations may not be available due to delayed feedback. Let $\Xbt_t=\{X_s\in\Xb_t: s+\tau_s\le t\}$ be the set of observation points for which the feedback has arrived by time $t$, and note that this is a random set of observations due to the stochastic delays. Simplifying the notation, we use $\mu_t,\sigma_t$ for the predictor and uncertainty estimate using $\Xb_t,\yb_t$, and  $\mut_t,\sigmat_t$ for the predictor and uncertainty estimate using $\Xbt_t,\ybt_t$, where $\ybt_t=[f(X_s)+\epsilon_s]^{\top}_{X_s\in \Xbt_t}$ is the vector of available observations. We note that in the delayed feedback setting, we can compute $\sigma_t$ in addition to $\mut_t,\sigmat_t$ since this only depends on the chosen inputs, not the observations. On the other hand, $\mu_t$ cannot be computed since it depends on some of the missing feedback. All three available statistics $\mut_t,\sigmat_t, \sigma_t$ are used in designing our algorithm, BPE-Delay.



\section{Batch Pure Exploration with Delays}\label{sec:algorithm}

In this section, we describe our proposed algorithm: Batch Pure Exploration with Delays (BPE-Delay). 
The algorithm proceeds in rounds $r=1,2,\dots, R$, where $R$ is the total number of rounds. Each round consists of $t_r$ time steps so that $\sum_{r=1}^Rt_r = T$. During each round, the observation points are selected based on a maximum uncertainty acquisition function (defined in~\eqref{eq:maxvar}). At the end of each round, confidence intervals are used to remove the points which are unlikely to be the maximiser of $f$.

The length $t_r$ of round $r$ is increasing in $r$ and is chosen carefully to balance the exploration needed in each round, taking into account the stochastically delayed feedback, and the number of rounds. In particular we set $t_r=\lceil q_r+u_{T}(\delta)\rceil$ where $u_{T}(\delta)$ is a $1-\delta$ upper bound on the delay random variable, and $q_r=\sqrt{q_{r-1}T}$ is determined recursively, for $r\ge 1$, initialized at $q_0=1$. The delay related quantity is set to $u_T(\delta)=\E[\tau]+\psi_{T}(\frac{\delta}{2})$ with 
\begin{equation}\label{psi}
    \psi_t(\delta) = \min\left\{\sqrt{2\xi^2\log(\frac{3t}{2\delta})},~2b\log(\frac{3t}{2\delta})\right\},
\end{equation}
where $\xi$ and $b$ are the parameters specified in Assumption~\ref{ass:sub-exp}. In the analysis in Section~\ref{Sec:Analysis}, we will see that $u_T(\delta)$ is a $1-\frac{\delta}{2}$ upper confidence bound on the delay random variable.
It turns out that this choice of $t_r$ depending on $u_T(\delta)$ is crucial in enabling us to improve the delay dependence of our algorithm.
If we know there is no delay in the observations, we can set $u_T(\delta)$ to zero. The length of the last round can also be easily adjusted to ensure $\sum_{t=1}^R t_r=T$.

BPE-Delay maintains a set $\Xc_r$ of potential maximisers  of~$f$. This set is recursively pruned from $\Xc_{r-1}$ using confidence intervals around each $x \in \Xc_{r-1}$ to remove those that are sub-optimal with high probability. We start with the full input space, $\Xc_0=\Xc$.
During each round $r$, the $k$-th observation point $X_{k,r}$ is selected from $\Xc_r$ based on a maximum uncertainty acquisition:
\begin{equation}\label{eq:maxvar}
X_{k,r}\gets\arg\max_{x\in\Xc_r}\sigma_{k-1,r}(x), 
\end{equation}
where $\sigma_{k,r}(\cdot)$ denotes the uncertainty estimate in \eqref{GP_reg} calculated using the points $\Xb_{k,r}=\{X_{1,r}, X_{2,r},\dots,X_{k,r}\}$. 
BPE-Delay uses only the inputs chosen in round $r$ so far, to calculate the acquisition function and choose the remainder of the points in round $r$ based on the uncertainty estimates. While using entire past observations may be effective, establishing corresponding regret bounds becomes difficult. The reason is that creating such dependency among observation points invalidates tight confidence intervals given in Lemma~\ref{lem:conf-single}. Nonetheless, \citep{li2022gaussian} showed in experiments that discarding the data collected in previous rounds, although may seem wasteful, does not significantly harm the performance. 
From definition of the uncertainty estimate given in~\eqref{GP_reg}, we can see that it does not depend on the observation values. Thus, $\sigma^2_{k,r}(\cdot)$ is well defined, despite not all observation values being available due to delay. We use double index $r=1,2,\dots$, $k=1,2,\dots, t_r$ for clarity of presentation. 
The observation points however can be indexed using $t=1,2,\dots$ by concatenating the observation points in all rounds (see Algorithm~\ref{alg:cap}).

\begin{algorithm}[h]
\caption{Batch Pure Exploration with Delays (BPE-Delay)}\label{alg:cap}
\begin{algorithmic}
\State Input: Action set $\Xc$, number of steps $T$;
\State Initialize: $\Xc_1\gets\Xc$, $q_0=1$, $t=1$ ;
\newline~~~~~\texttt{\textcolor{blue}{\# Algorithm proceeds in rounds $r=1,2,\dots$.}}
\For{$r=1,2, \dots$, until $t\le T$}
\State $q_r=\lceil\sqrt{Tq_{r-1}}\rceil$ 
~~~~~\texttt{\textcolor{blue}{\# $q_r$ is used to determine $t_r$}}
\State $t_r=\lceil q_r+u_T(\delta)\rceil$ ~~~~~\texttt{\textcolor{blue}{\# $t_r$ is the length of round $r$}}
\For{$k=1,\dots,t_r$}
\State 
\texttt{\textcolor{blue}{\# the points in each round are chosen by a maximum uncertainty acquisition function}}
\newline $X_{k,r}\gets\arg\max_{x\in\Xc_r}\sigma_{k-1,r}(x)$  
\newline
$X_t\gets X_{k,r}$
\newline 
$t=t+1$
\EndFor
\newline
~~~~~\texttt{\textcolor{blue}{\# Remove the input points which are unlikely to be the maximiser of $f$,  using confidence intervals}}
\newline $\Xc_{r+1}\gets\{x\in\Xc_r: U_{r,\delta}(x)\ge L_{r,\delta}(z), \forall z\in\Xc_r \}$  
\EndFor

\end{algorithmic}
\end{algorithm}

To contrast the statistics derived from the entire set of observations in round $r$ ---ignoring the delay---, and the ones using available observations ---considering the delay---, we recall the notations $\mut_{k,r}$ and $\sigmat_{k,r}$ for the statistics using just the available observations. Specifically, $\mut_{k,r}$ and $\sigmat_{k,r}$ are the prediction and uncertainty estimate after selecting $k$ observation points in round $r$ using the set $\Xbt_{k,r}= \{X_{s,r}\in \Xb_{k,r}:s+\tau_{s,r}\le  k\}$ of points selected in round $r$, with available observations $\ybt_{k,r}=[y=f(X_{s,r})+\epsilon_{s,r}]_{X_{s,r}\in\Xbt_{k,r}}$ in round $r$.

At the end of each round $r$, using the available observations in round $r$, we create upper and lower  confidence bounds on $f(\cdot)$, $U_r(\cdot)$ and $L_r(\cdot)$ respectively. These are used to remove points which are unlikely to be a maximiser of $f$. In particular, if there exists a point $z\in\Xc_r$ for which the lower confidence bound is larger than the upper confidence bound for $x\in \Xc_r$, then $x$ is unlikely to be the maximiser of $f$ so we can remove it. The confidence interval widths are selected in a way that all the confidence intervals used in the algorithm hold true with high probability (see Theorems~\ref{the:regret} and~\ref{the:regretcont} for details). 


We emphasize that while we use $\sigma_{k,r}$ for selecting the observation points within each round, $\mut_{k,r}$ and $\sigmat_{k,r}$ are used for forming the confidence intervals at the end of each round. Using $\sigma_{k,r}$ avoids selecting repetitive points due to unavailable feedback, while the valid confidence intervals can only be formed using $\mut_{k,r}$ and $\sigmat_{k,r}$.

 The complete pseudo-code for the BPE-Delay algorithm is given in Algorithm~\ref{alg:cap}. 






\section{Analysis}
\label{Sec:Analysis}





In this section, we provide the analysis of the BPE-Delay algorithm. Specifically, we prove a high probability $\Oct(\sqrt{T\Gamma_k(T)} + \E[\tau])$ bound on its regret. This regret bound completely decouples the effect of the delay from the dominant time dependent regret term. We note that the additional regret due to delay in our result is significantly smaller than the existing work and matches what is seen in the simpler $K$-armed bandit problem \citep{joulani2013online}. 

We consider two cases of finite and continuous $\Xc$ separately, due to minor differences in the confidence intervals which lead to mild differences in the regret bounds. In particular, when $\Xc$ is finite, the regret bound scales with $\Oc(\sqrt{\log(|\Xc|)})$. In the case of a continuous and compact $\Xc\subset\Rr^d$, under Assumption \ref{ass:disc}, we prove a similar regret bound which scales with an $\Oc(\sqrt{d})$ factor instead of $\Oc(\sqrt{\log(|\Xc|)})$. While in the kernel bandit setting, $\sqrt{d}$ is typically hidden in the $\Oc$ notation as a constant not growing with $T$, in the linear bandit setting, the dependency on $d$ should be pronounced and is important in evaluating the algorithms.

\begin{theorem}\label{the:regret}
Consider the kernel based bandit problem with delayed feedback described in Section~\ref{sec:PF}.
When $\Xc$ is finite, set the confidence intervals in BPE-Delay (Algorithm~\ref{alg:cap}) to
\begin{eqnarray}\nn
    U_{r,\delta}(x) = \mut_{t_r,r}(x)+\betat(\delta)\sigmat_{t_r,r}(x),\\
    L_{r,\delta}(x) = \mut_{t_r,r}(x)-\betat(\delta)\sigmat_{t_r,r}(x),
\end{eqnarray}
with $\betat(\delta) = C_k+\frac{\sigma}{\lambda}\sqrt{2\log(\frac{4R|\Xc|}{\delta})}$.
Under Assumptions~\ref{ass:f_norm},~\ref{ass:noise},~\ref{ass:sub-exp}, with probability at least $1-\delta$, the regret of BPE-Delay is bounded by
\begin{equation}
    \Rc(T) = \Oc\left(\betat(\delta)\log\log(T)\sqrt{T\Gamma_k(T)} + \E[\tau]\right).
\end{equation}

\end{theorem}

Note that with a finite $\Xc$, $\betat(\delta)=\Oc(1)$, not depending on problem parameters such as $d$ and $T$. In the next theorem, we bound the regret when $\Xc$ is not finite, but a compact subset of $\Rr^d$.

\begin{theorem}\label{the:regretcont}
Consider the kernel based bandit problem with delayed feedback described in Section~\ref{sec:PF}, under Assumptions~\ref{ass:f_norm},~\ref{ass:noise},~\ref{ass:sub-exp} and~\ref{ass:disc}. 
Set the confidence intervals in BPE-Delay (Algorithm~\ref{alg:cap}) to
\begin{equation}\nn
    U_{r,\delta}(x) = \mut_{t_r, r}(x)+\frac{2}{u_r}+\betat'_{r}(\delta)(\sigmat_{t_r,r}(x)+\frac{2}{\sqrt{u_r}}),
\end{equation}
\begin{equation}
    U_{r,\delta}(x) = \mut_{t_r, r}(x)+\frac{2}{u_r}+\betat'_{r}(\delta)(\sigmat_{t_r,r}(x)+\frac{2}{\sqrt{u_r}}),
\end{equation}
with $\betat_r'(\delta)=\beta'_{u_r}(\frac{\delta}{4R})$ and $\beta'_t$ given in Lemma~\ref{lem:conf-unif}.
Then, with probability at least $1-\delta$, the regret of BPE-Delay is bounded by,
\begin{equation}
    \Rc(T) = \Oc\left(\betat'_R(\delta)\log\log(T)\sqrt{T\Gamma_k(T)} + \E[\tau]\right).
\end{equation}

\end{theorem}

From Lemma~\ref{lem:conf-unif}, we can see that $\betat'_R(\delta) = \Oc(C_k+\frac{\sigma}{\lambda}\sqrt{d\log(\frac{TR C_k}{\delta})})$. 
Thus the regret bound scales with $\Oc(\sqrt{d})$ that becomes particularly important when applied to linear bandits.

\begin{proof}[Proof Sketch]
By the choice of size $t_r$ of each round, the number $R$ of rounds is bounded as $R= \Oc(\log\log(T))$. 
We first show that the uncertainty estimate using $t_r$ observations $\sigmat_{t_r,r}$ sufficiently shrinks.
Using the confidence intervals, we prove that i) the maximiser $x^*$ is not removed during any round; i.e. $x^*\in\Xc_r$, for all $r$, and ii) 
the instantaneous regret incurred by selecting each point $X_{k,r}\in \Xc_r$ is bounded. The bound on $f(x^*)-f(X_{k,r})$ is based on the amount of exploration in previous round, taking into account the delay. Using the bound on instantaneous regret and the size $t_r$ of round $r$ we bound the regret in round $r$. A detailed proof is provided in the appendix. 
\end{proof}

Theorem~\ref{the:regret} can be specialized for many commonly used kernels including Mat{\'e}rn and SE, as well as for the special case of linear kernels.  

\begin{corollary}\label{cor:spec}
Under the setting of Theorem~\ref{the:regretcont}, the following hold with probability at least $1-\delta$,
\begin{itemize}
\item In the case of Mat\'ern kernel with smoothness parameter $\nu$, 
\begin{eqnarray}
\Rc(T)=\Oct\left(T^{\frac{\nu+d}{2\nu+d}} +\E[\tau]\right).
\end{eqnarray}
\item In the case of SE kernel, 
\begin{eqnarray}
\Rc(T)=\Oc\left(\sqrt{T}(\log(T))^{d+1}\log\log(T) + \E[\tau]\right).
\end{eqnarray}
\item In the case of linear kernel, 
\begin{eqnarray}
\Rc(T)=\Oc\left(d\sqrt{T}\log(T)\log\log(T)+\E[\tau]\right).
\end{eqnarray}
\end{itemize}

\end{corollary}

Comparing to the lower bound in the case of linear bandit~\citep{lattimore2020bandit}, and the lower bounds for kernel bandit in the case of SE and Mat{\'e}rn kernels~\citep{Scarlett2017Lower}, the regret bounds shown in Corollary~\ref{cor:spec} are order optimal, up to logarithmic factors and the additive delay penalty which is independent of all other problem parameters. We thus reduced the delay related term in the regret bounds to only $\E[\tau]$ without affecting the dominant term in the regret bound. For comparison, in the (generalized) linear setting the best known bound on the delay related regret was $\Oc(d^{\frac{3}{2}}\E[\tau])$~\citep{howson2022delayed}; that is improved with a $d^{\frac{3}{2}}$ factor in our results. In the kernel bandit setting, we also improved the $\Oc(\Gamma_T\E[\tau])$ delay related regret~\citep{verma2022bayesian} to only $\Oc(\E[\tau])$. This represents a significant improvement considering that $\Gamma_T$ can become arbitrarily close to linear in $T$, in the case of a Mat{\'e}rn kernel with a small $\nu$ and large $d$. This is in addition to the improvement in the first term in regret from $\Oct(\Gamma_T\sqrt{T})$ to $\Oct(\sqrt{\Gamma_T T})$.

\begin{figure}[t]
    \centering
    \subfloat[\centering $f_1$]{{\includegraphics[width=3.5cm, height=2.80cm]{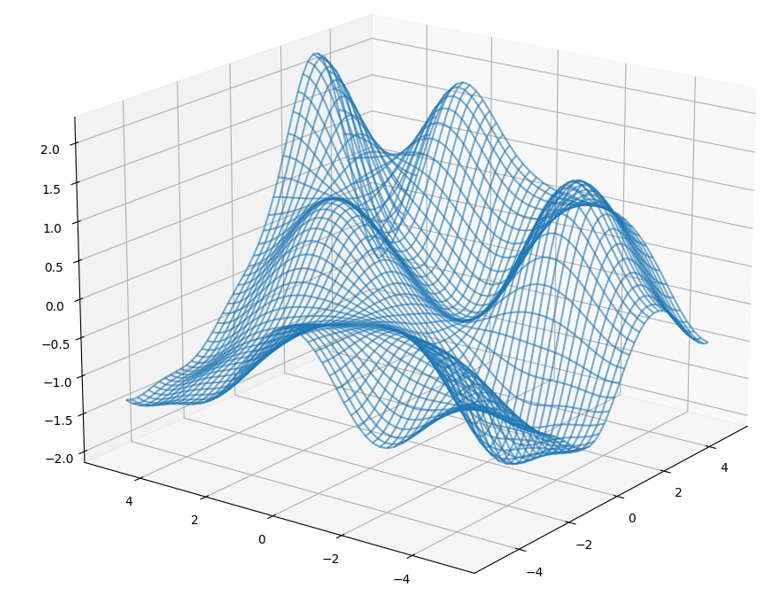} }}%
    \qquad
    \subfloat[\centering $f_2$]{{\includegraphics[width=3.5cm, height=2.80cm]{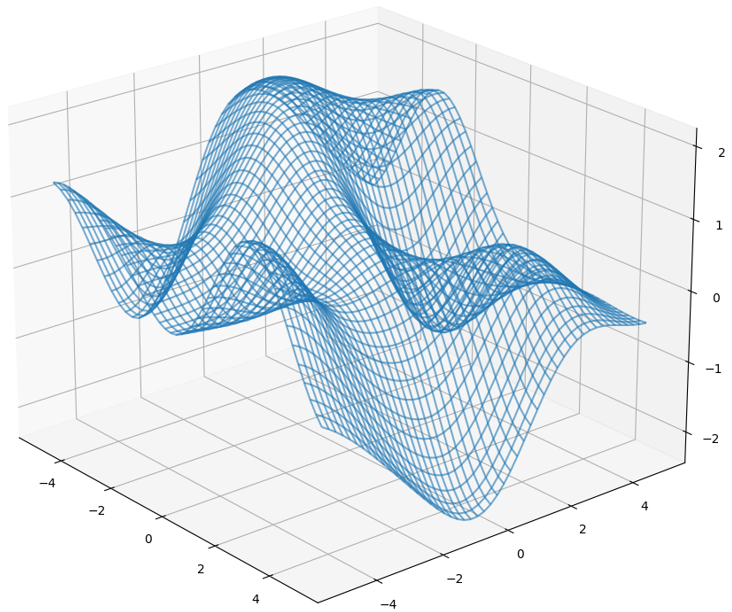} }}%
    \caption{Objective functions used in our experiments.}%
    \label{Fig:objectivef}%
\end{figure}

\section{Experiments}\label{sec:exp}

Following our theoretical analysis, we provide numerical experiments on the performance of BPE-Delay and compare it to the GP-UCB-SDF~\citep{verma2022bayesian}. We test the algorithms on the objective functions $f_1$ and $f_2$ shown in Figure~\ref{Fig:objectivef}. These functions are generated by fitting a kernel based model to points randomly generated from a multivariate Gaussian. This is a common technique to create RKHS elements~\citep[e.g., see][]{Chowdhury2017bandit, vakili2021optimal, li2022gaussian}. We use a SE kernel with a length scale parameter $l = 0.8$ for $f_1$ and $l = 1.0$ for $f_2$ in order to generate these objective functions. 
The learner can then choose from $|\mathcal{X}| = 2500$ points over a uniform $50 \times 50$ grid. The sampling noise is zero mean Gaussian with standard deviation $\sigma=0.02$. The stochastic delay in the feedback is generated from a Poisson distribution with parameter $\lambda$. 
The calculation of $\psi_{t}$ for BPE-Delay uses $\xi = 9$ and $b = 1$ given in Assumption~\ref{ass:disc}. The cumulative regret curves are the average of $10$ independent experiments. The error bars indicate half a standard deviation. The code for these experiments is provided in an anonymous GitHub repository (see the supplementary material). 

\begin{figure}[!htb]
    \centering
    \subfloat[\centering $f_1$]{{\includegraphics[width=6cm, height=2.75cm]{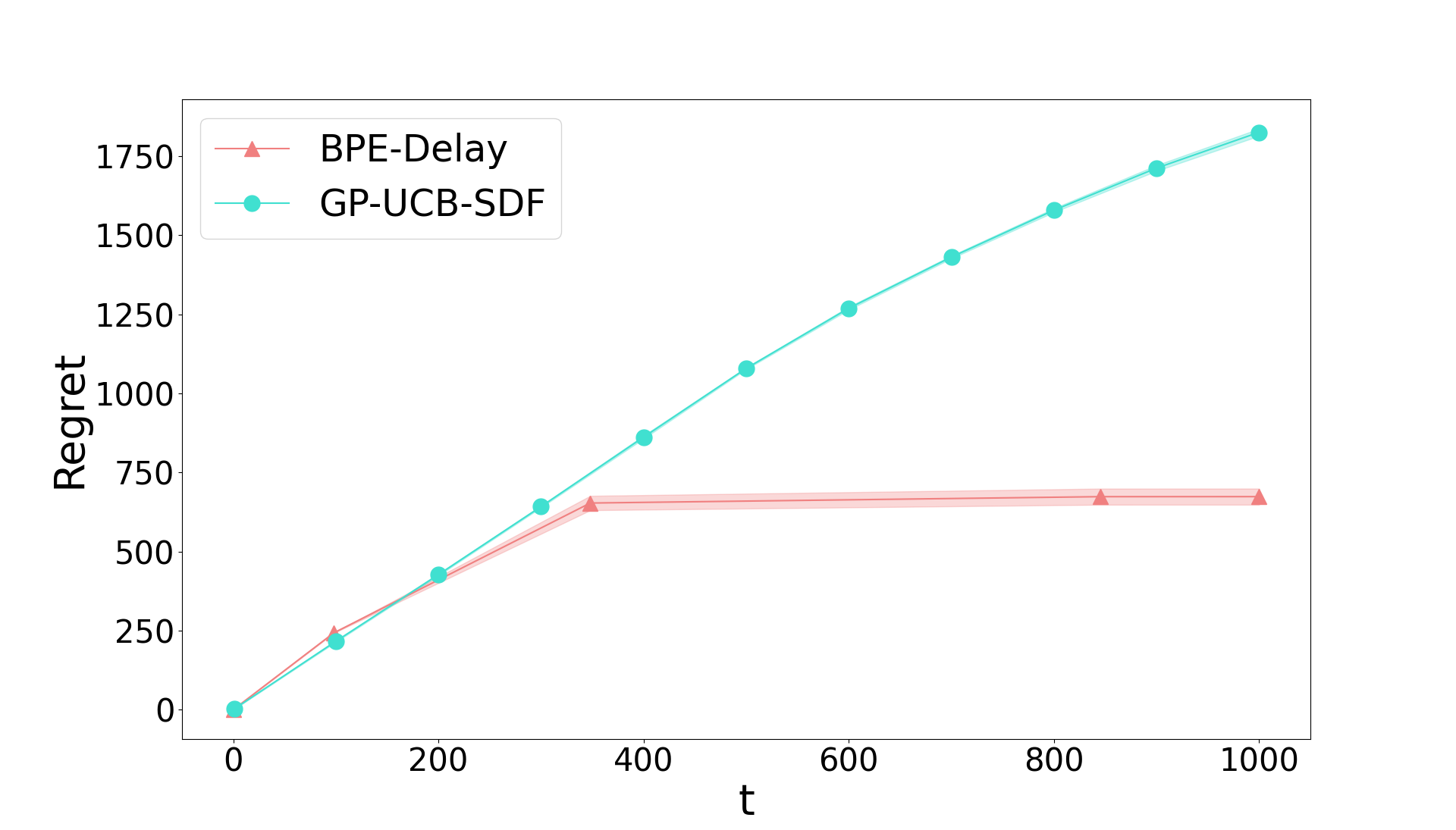} }}%
    \qquad
    \subfloat[\centering $f_2$]{{\includegraphics[width=6cm, height=2.75cm]{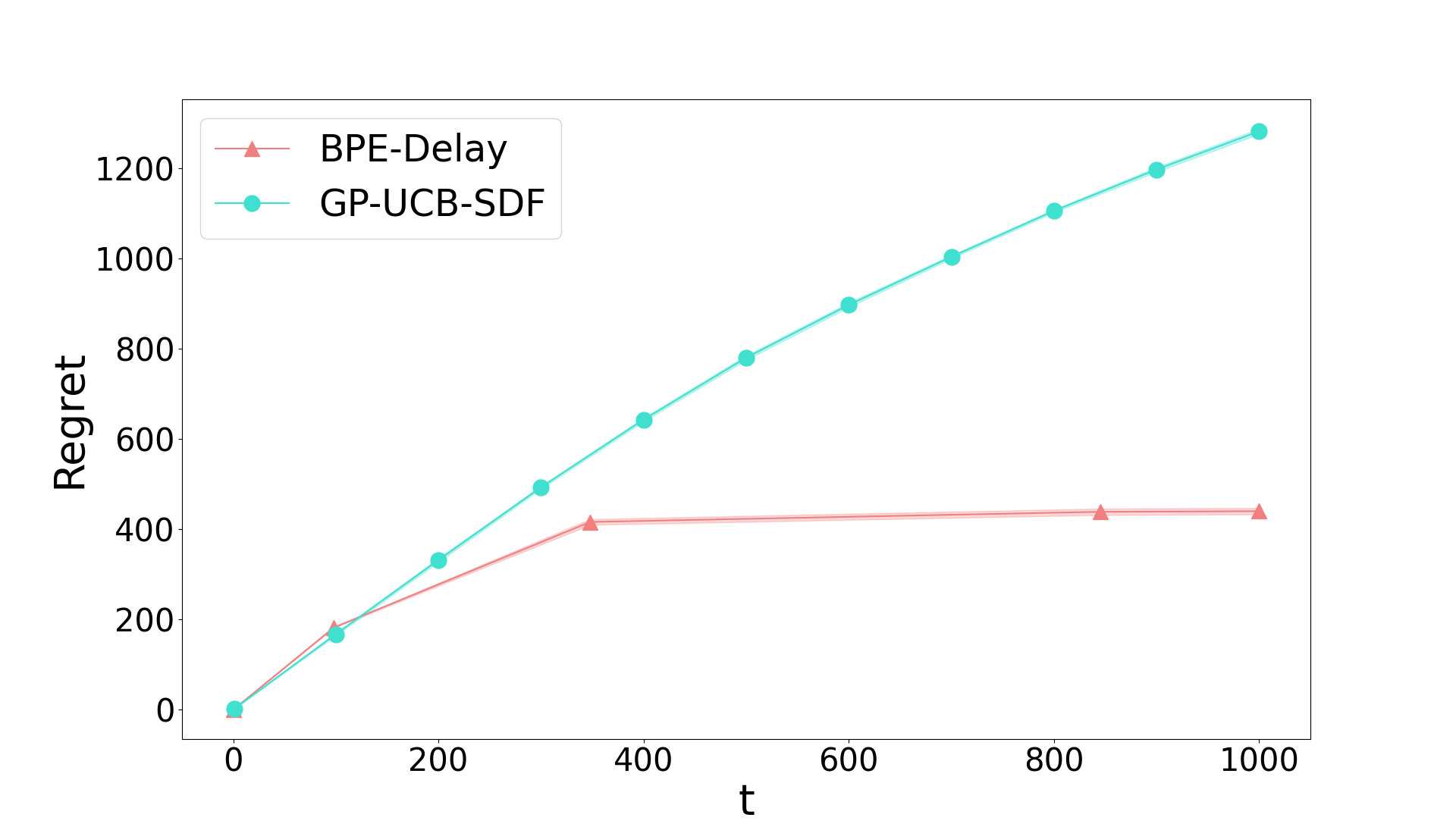} }}%
    \caption{Comparing regret performance of BPE-Delay and GP-UCB-SDF.}%
    \label{Fig:comaprewithSDF}
\end{figure}

In Figure~\ref{Fig:comaprewithSDF}, we compare the regret performance of BPE-Delay with GP-UCB-SDF introduced in the most related work in the same setting as ours~\citep{verma2022bayesian}. The figure shows a significant improvement in the regret performance using BPE-Delay in both cases. In this experiment, the delays are Poisson distributed with $\lambda=50$. 

\begin{figure}[!htb]%
    \centering
    \subfloat[\centering $f_1$]{{\includegraphics[width=6cm, height=2.75cm]{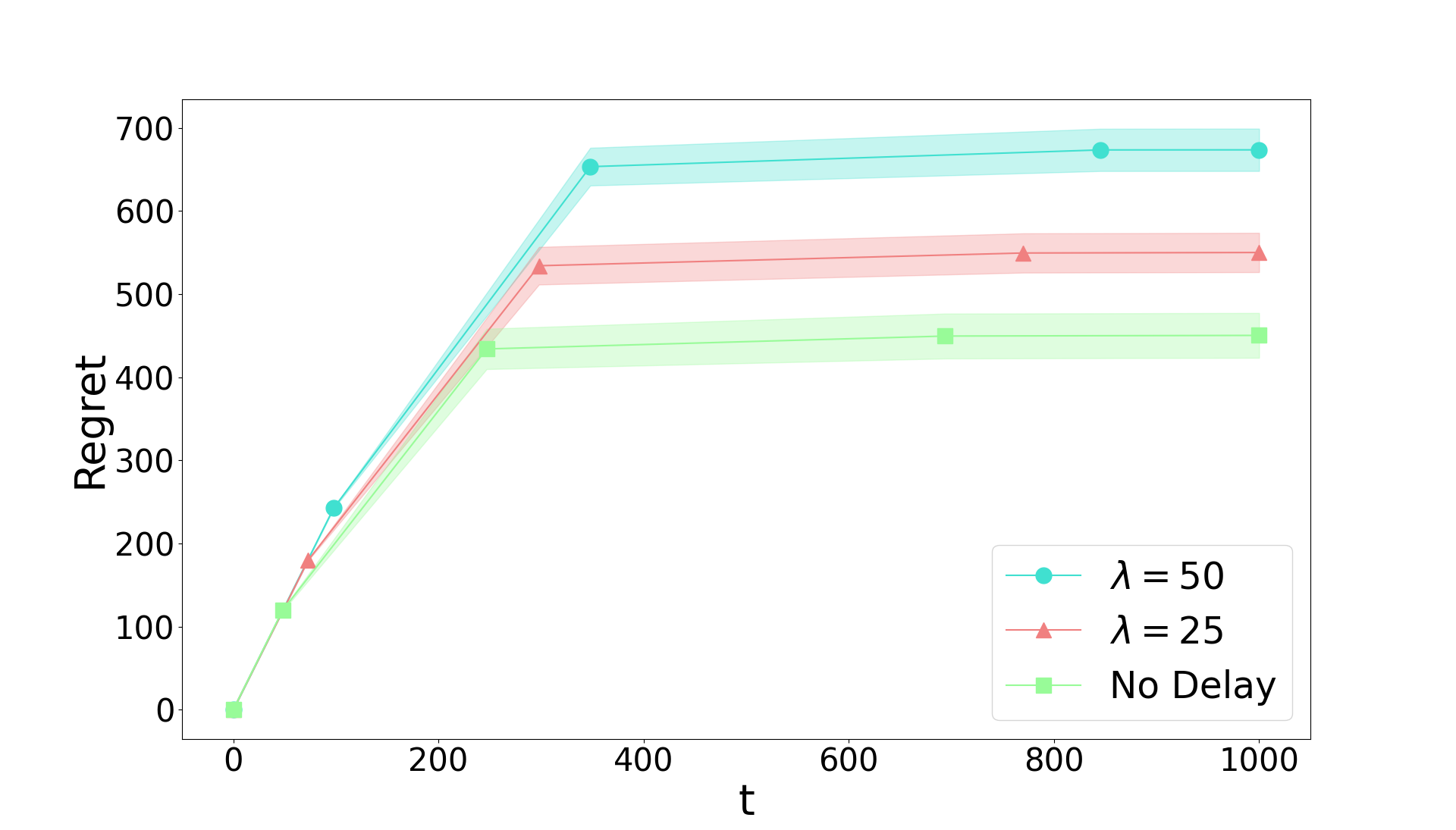} }}%
    \qquad
    \subfloat[\centering $f_2$]{{\includegraphics[width=6cm, height=2.75cm]{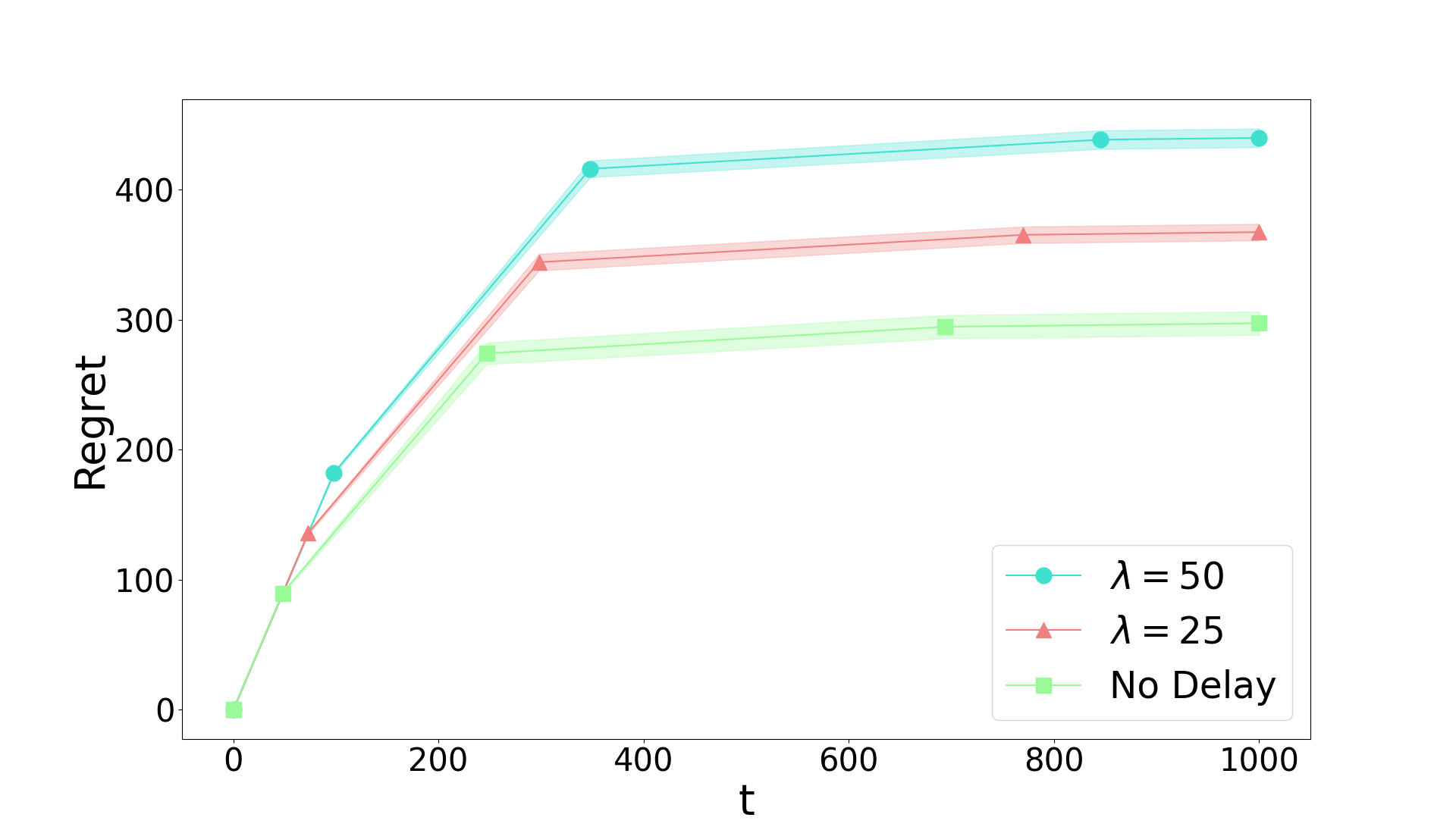} }}%
    \caption{Regret performance of BPE-Delay for varying delay Poisson parameters.}%
    \label{Fig:lambdaDelya}%
\end{figure}

\begin{figure}[!htb]%
    \centering
    \subfloat[\centering $f_1$]{{\includegraphics[width=6cm, height=2.75cm]{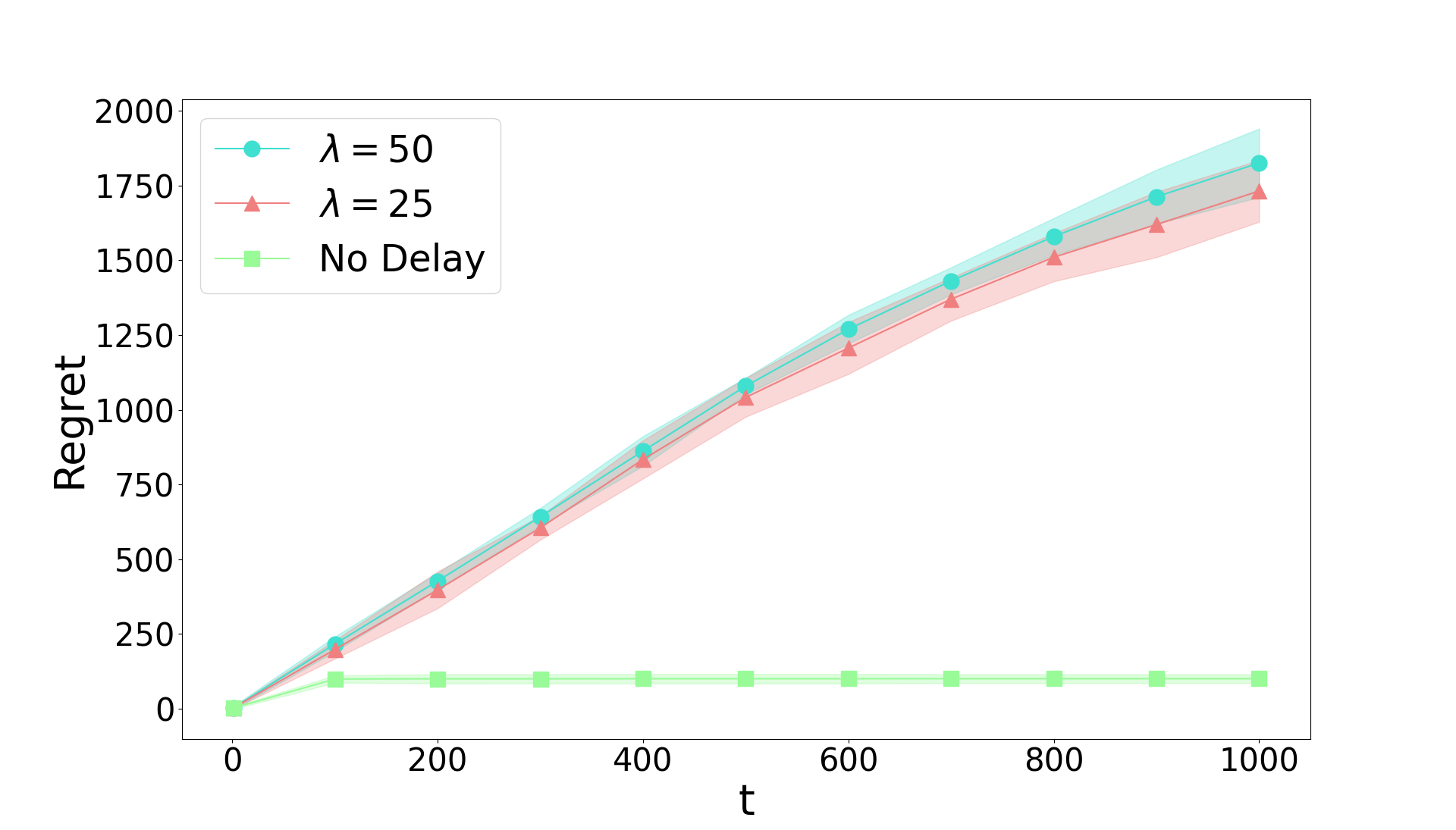} }}%
    \qquad
    \subfloat[\centering $f_2$]{{\includegraphics[width=6cm, height=2.75cm]{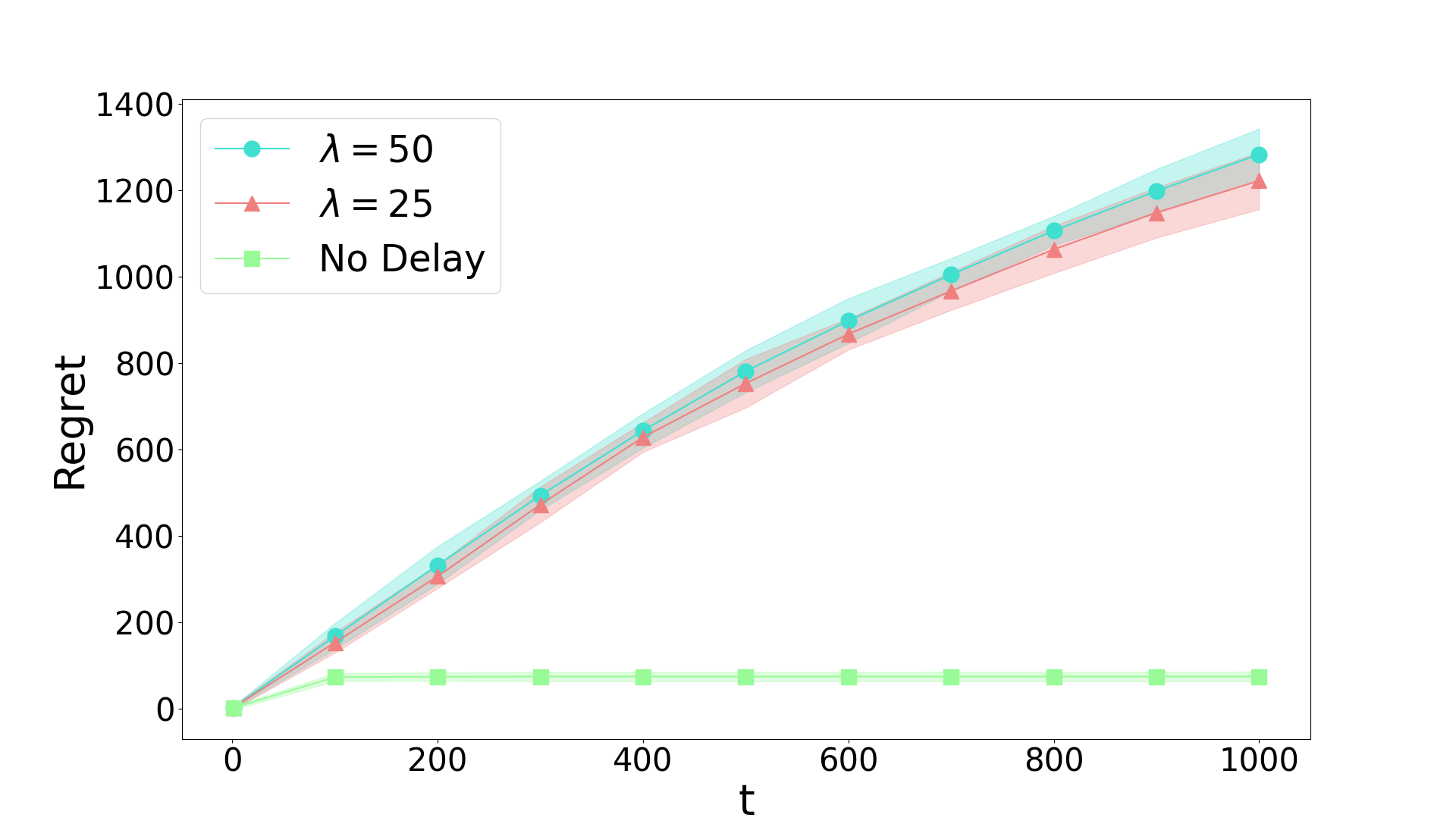} }}%
    \caption{Regret performance of GP-UCB-SDF for varying delay Poisson parameters.}%
    \label{Fig:lambdaSDF}%
\end{figure}

In Figures~\ref{Fig:lambdaDelya} and~\ref{Fig:lambdaSDF}, we show the effect of delay for these two algorithms. Specifically, we vary the Poisson delay parameter as $\lambda=0,25$ and $50$, while maintaining the rest of parameters as the previous experiment. BPE-Delay shows a linear excess in the regret with the expected delay. GP-UCB-SDF on the other hand shows dramatic jump in regret even with moderate delay values. 

\begin{figure}[!htb]
    \centering
    \subfloat[\centering $f_1$]{{\includegraphics[width=6cm, height=2.75cm]{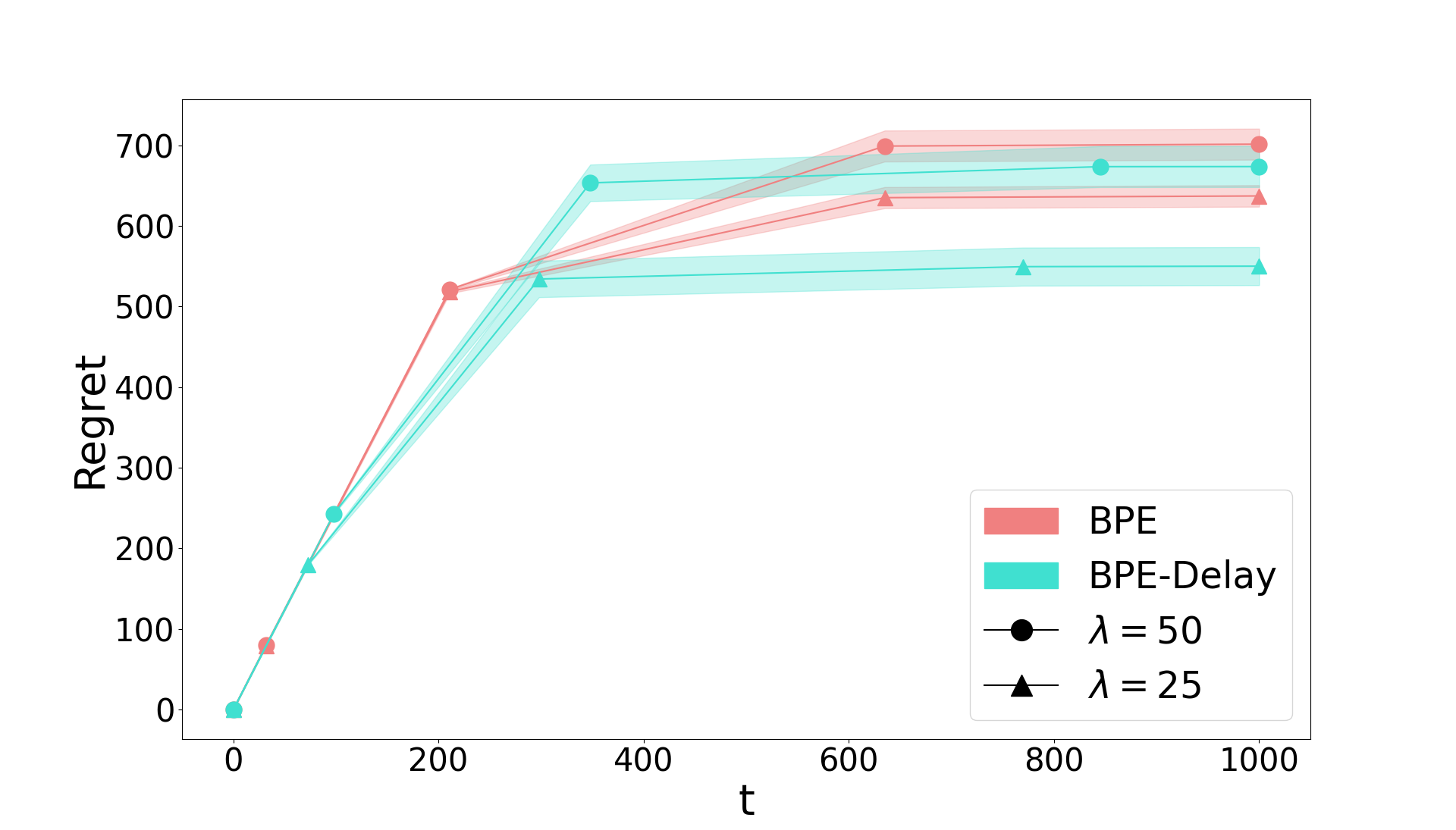} }}%
    \qquad
    \subfloat[\centering $f_2$]{{\includegraphics[width=6cm, height=2.75cm]{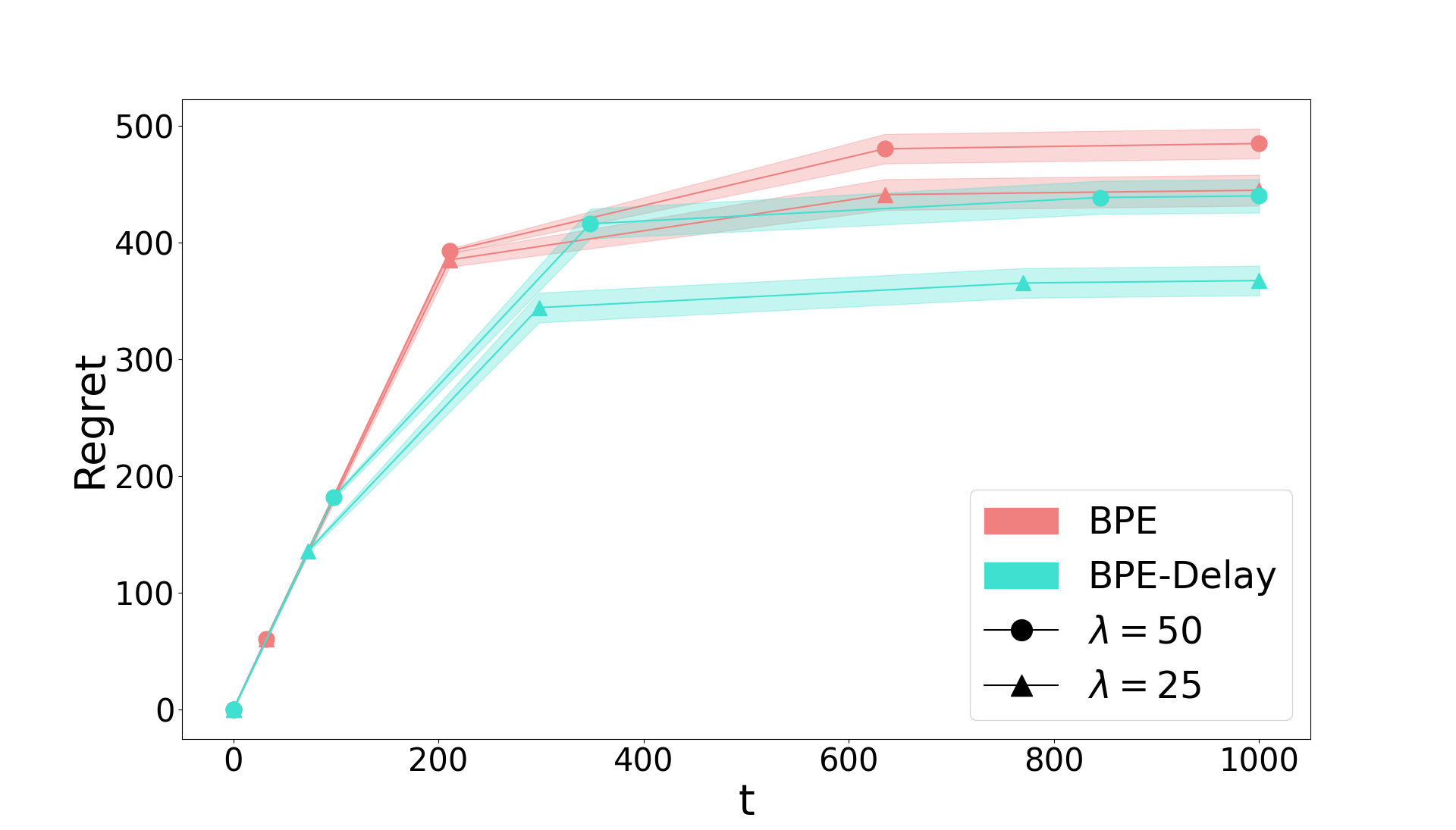} }}%
    \caption{Comparing the regret performance of BPE and BPE-Delay in the delayed feedback setting.}%
    \label{Fig:5}%
\end{figure}

We also compare the performance of the BPE-Delay and BPE~\citep{li2022gaussian}. As we can see from Figure~\ref{Fig:5}, BPE-Delay naturally performs better than BPE in a delayed setting.


\section{Conclusion}

There has been a great attention paid to kernel bandits due to their numerous applications in machine learning, and academic and industrial experimental design. In many real world scenarios, e.g., recommender systems, the feedback from the decisions are not immediately available, naturally leading to the formulation of kernel bandits with stochastically delayed feedback. For this setting, we proposed BPE-Delay which is able to efficiently deal with the delayed feedback. We showed an order optimal regret bound for BPE-Delay with a very small excess in regret due to the delayed feedback, significantly improving upon the existing work in the same setting. We also show that these theoretical improvements are maintained in several simulation studies. In addition, when applied to the special case of linear kernels, our theoretical regret bounds improve the delay related penalty by a factor of $d^{\frac{3}{2}}$ compared to the state of the art. In all cases, our results are the first to show that the additive delay penalty only depends on $\E[\tau]$.
This extends what is known for the simpler $K$-armed bandit setting to the much more realistic kernel bandit setting.


\bibliography{references.bib}
\bibliographystyle{abbrvnat}

\appendix
\section{Proofs}
In this section, we provide detailed proofs for Theorems~\ref{the:regret},~\ref{the:regretcont} and Corollary~\ref{cor:spec}. 
We structure the proof of theorems as follows. We first overview the kernel based confidence intervals, the bounds on the delay and the number of rounds. After formalizing these bounds, we proceed with the proof in three steps. In the first step, we bound the maximum uncertainty at any point $x\in\Xc_{r+1}$ at the end of each round $r$. In the second step, we bound the instantaneous regret for selecting a point in each round. In the third step, we bound the total regret. At the end of the proof, we discuss Corollary~\ref{cor:spec}.

\paragraph{Confidence Intervals.}
Recall the confidence intervals used in the BPE-Delay algorithm to update the set $\Xc_r$ of potential maximisers in round $r$. Using Lemmas~\ref{lem:conf-single} and~\ref{lem:conf-unif} and a probability union bound over rounds, all confidence intervals used in the algorithm are valid with probability at least $1-\frac{\delta}{2}$. Specifically, let us define the event
\begin{eqnarray}\label{def:E}
\Ec = \{L_{r,\delta}(x)\le f(x)\le U_{r,\delta}(x), \forall r=1,2,\dots, R, \forall x\in\Xc_r\}.
\end{eqnarray}

Then, we have $\Pr[\Ec]\ge 1-\frac{\delta}{2}$. 

\paragraph{Delay.}

Recall the definition of~$\psi_t(\delta)$ given in~\eqref{psi}. We have the following concentration for sub-exponential random variables.

\begin{lemma}[\cite{howson2022delayed}]\label{lem:subexp}
Let $\tau_t-\E[\tau_t]$, $t=1,2,\dots$, be i.i.d. sub-exponential random variables with parameters $\xi$ and $b$ as specified in Assumption~\ref{ass:sub-exp}. 
Then, 
\begin{eqnarray}
    \Pr\left[\forall t\ge 1, \tau_t\le \E[\tau]+\psi_t(\delta)\right]\ge 1-\delta.
\end{eqnarray}
\end{lemma}

Let $\tau_{k,r}$ denote the random delay for the $k$-th observation in round $r$ of BPE-Delay.
Let us define the event 
\begin{eqnarray}\label{def:Eprime}
\Ec'=\{\forall r, \forall k\le q_r: \tau_{k,r}\le u_T(\delta) \}.
\end{eqnarray}
Using Lemma~\ref{lem:subexp} and definition of $u_T$, we have $\Pr[\Ec']\ge 1-\frac{\delta}{2}$. 

We thus have the probability that both $\Ec$ and $\Ec'$ hold true $\Pr[\Ec\cap \Ec']\ge 1-\delta$. 

\textbf{We condition the rest of the proof on} $\Ec\cap \Ec'$.

\paragraph{The number of rounds.}
The number $R$ of rounds in BPE-Delay is bounded in the following lemma. 
\begin{lemma}
    Recall the choice of round lengths $t_r$ in BPE-Delay. We have $R=\Oc(\log\log(T))$.
\end{lemma}
The proof follows from similar steps as in the proof of Proposition $1$ in~\cite{li2022gaussian}. 

We now proceed to the main steps in the proof of theorems.

\paragraph{Step 1 (Maximum uncertainty at the end of each round).}
In each round $r$,
the sum of uncertainties using all observations (including the ones with delayed feedback) at the end of $q_r$ observations can be bounded using $\Gamma_k(q_r)$. 

\begin{lemma}\label{lem:sriv}[\cite{srinivas2010gaussian}]
Recall definition of $\Gamma_k(t)$ given in \eqref{def:gamma}, for a set of $t$ observation $\Xb_t$, we have 
\begin{eqnarray}
\sum_{s=1}^t\sigma^2_{\Xb_{s-1}}(X_s)\le c_1\Gamma_k(t), 
\end{eqnarray}
where $c_1=\frac{2}{\log(1+\frac{1}{\lambda^2})}$. 
\end{lemma}
Thus, considering the observation points in round $r$, we have
\begin{eqnarray}\label{eq:varsum}
\sum_{k=1}^{q_r}\sigma^2_{k-1,r}(X_{k,r}) \le c_1\Gamma_{k}(q_r),
\end{eqnarray}
where $c_1$ is the constant given in Lemma~\ref{lem:sriv}. By the selection rule of observation points based on maximum uncertainty acquisition~\eqref{eq:maxvar}, we have, for all $x\in\Xc_r$, $1\le k\le q_r$
\begin{eqnarray}
    \sigma_{k-1,r}(x) \le \sigma_{k-1,r}(X_{k,r}).
\end{eqnarray}
In addition, by positive definiteness of the kernel matrix, conditioning on a bigger set of observations reduces the uncertainty (see \eqref{GP_reg}). Thus, 
\begin{eqnarray}
    \sigma_{q_r,r}(x) \le \sigma_{k-1,r}(X_{k,r}).
\end{eqnarray}

Combining the last inequality with~\eqref{eq:varsum}, we obtain
\begin{eqnarray}
    \sum_{k=1}^{q_r}\sigma^2_{q_r,r}(x) \le c_1\Gamma_{k}(q_r),
\end{eqnarray}
that implies
\begin{eqnarray}
\sigma^{2}_{q_r,r}(x)\le \frac{c_1\Gamma_{k}(q_r)}{q_r}.
\end{eqnarray}

Recall the event $\Ec'$ given in~\eqref{def:Eprime}.
Under $\Ec'$, $\sigmat^2_{t_r,r}$ is conditioned on a larger set of observations compared to $\sigma^2_{q_r,r}$. The positive definiteness of the kernel matrix implies that conditioning on a larger set of observations reduces the uncertainty.
Thus, we have
$\sigmat^2_{t_r,r}(x)\le \sigma^2_{q_r,r}(x), \forall x\in\Xc_r $. As a result,
\begin{eqnarray}
\sigmat^2_{t_r,r}(x)\le \frac{c_1\Gamma_{k}(q_r)}{q_r}, ~~~\forall x\in\Xc_r.
\end{eqnarray}

\paragraph{Step 2 (Instantaneous regret).}

We can use the confidence intervals for RKHS elements given in Lemmas~\ref{lem:conf-single} and~\ref{lem:conf-unif} to bound the instantaneous regret as follows. Recall the event $\Ec$ defined in~\eqref{def:E}.

We first note that for all $x\in\Xc_r$,
\begin{eqnarray}\nn
U_{r,\delta}(x^*)&\ge& f(x^*)\\\nn
&\ge& f(x)\\\nn
&\ge& L_{r,\delta}(x).
\end{eqnarray}
Thus, $U_{r,\delta}(x^*)\ge \max_{x\in\Xc_r}L_{r,\delta}(x)$ for all $r$, and $x^*$ will not be eliminated during any round of BPE-Delay. Therefore $x^*\in\Xc_r$, for all $r$.

We now proceed to bounding the instantaneous regret under two different settings of finite $\Xc$ and continuous $\Xc$, separately. The two cases correspond to Theorems~\ref{the:regret} and~\ref{the:regretcont}, respectively. 

\textcolor{blue}{\emph{When $\Xc$ is finite (The setting of Theorem~\ref{the:regret}):}}

For all $x\in \Xc_{r+1}$,

\begin{eqnarray}\nn
f(x^*) - f(x)&\le& U_{r,\delta}(x^*)-L_{r,\delta}(x)\\\nn
&=& \mut_{t_r,r}(x^*)+\betat(\delta)\sigmat_{t_r,r}(x^*) -\mut_{t_r,r}(x)-\betat(\delta)\sigmat_{t_r,r}(x^*)\\\nn
&=&L_{r,\delta}(x^*)-U_{r,\delta}(x) + 2\betat(\delta)\sigmat_{t_r,r}(x^*) + 2\betat(\delta)\sigmat_{t_r,r}(x)\\\label{eq:rule}
&\le& 2\betat(\delta)\sigmat_{t_r,r}(x^*) + 2\betat(\delta)\sigmat_{t_r,r}(x).
\end{eqnarray}

The first and second equality follow from the choice of the confidence intervals in Theorem~\ref{the:regret}. The last line follows from the choice of $\Xc_{r+1}$ that ensures for all $x,z\in\Xc_{r+1}$, $L_{r,\delta}(z)\le U_{r,\delta}(x)$. 

Combining with the bound on $\sigmat_{t_r,r}(x)$ obtained in the previous step, we bound the instantaneous regret for all $x\in\Xc_{r+1}$ as follows.
\begin{eqnarray}\label{eq:instdisc}
    f(x^*) -f(x)\le  4\betat(\delta)\sqrt{\frac{c_1\Gamma_k(q_r)}{q_r}}.
\end{eqnarray}

\textcolor{blue}{\emph{When $\Xc$ is continuous (The setting of Theorem~\ref{the:regretcont}):}}

Following similar steps as in the previous case, for all $x\in\Xc_{r+1}$,
\begin{eqnarray}\nn
f(x^*)-f(x)&\le& U_{r,\delta}(x^*)-L_{r,\delta}(x)\\\nn
&\le& \mut_{t_r, r}(x^*)+\frac{2}{q_r}+\betat'_{r}(\delta)(\sigmat_{t_r,r}(x^*)+\frac{2}{\sqrt{q_r}})\\\nn
&& ~~~- \mut_{t_r, r}(x)+\frac{2}{q_r}+\betat'_{r}(\delta)(\sigmat_{t_r,r}(x)+\frac{2}{\sqrt{q_r}})\\\nn
&=& L_{r,\delta}(x^*) - U_{r,\delta}(x)\\\nn
&&~~~+\frac{8}{q_r} + 2\betat'_{r}(\delta)\left(\sigmat_{t_r,r}(x^*)+\sigmat_{t_r,r}(x)+\frac{4}{\sqrt{q_r}}\right) \\\nn
&\le& \frac{8}{q_r} + 2\betat'_{r}(\delta)\left(\sigmat_{t_r,r}(x^*)+\sigmat_{t_r,r}(x)+\frac{4}{\sqrt{q_r}}\right).
\end{eqnarray}

Combining with the bound on $\sigmat_{t_r,r}(x)$ obtained in the previous step, we bound the instantaneous regret for all $x\in\Xc_{r+1}$ as follows.
\begin{eqnarray}\label{eq:instcont}
    f(x^*) -f(x)\le \frac{8}{q_r} + 4\betat'_{r}(\delta)\left(\sqrt{\frac{c_1\Gamma_k(q_r)}{q_r}}+\frac{2}{\sqrt{q_r}}\right).
\end{eqnarray}

\paragraph{Step 3 (Bounding cumulative regret):} 

Let $\Rc_r=\sum_{k=1}^{t_r}(f(x^*)-f(X_{k,r}))$ be the total regret in round $r$. Then, we have $\Rc(T)=\sum_{r=1}^R\Rc_r$. 
Summing up the regret in round $r$, we obtain the following.

\textcolor{blue}{\emph{When $\Xc$ is finite (The setting of Theorem~\ref{the:regret}):}}

Replacing the bound on instantaneous regret from~\eqref{eq:instdisc}, for $r\ge 2$,  
\begin{eqnarray}\nn
    \Rc_r&\le& 4t_{r}\betat(\delta)\sqrt{\frac{c_1\Gamma_k(q_{r-1})}{q_{r-1}}}\\\nn
    &\le& 4 \lceil\sqrt{Tq_{r-1}}+u_{T}(\delta)\rceil\betat(\delta)\sqrt{\frac{c_1\Gamma_k(q_{r-1})}{q_{r-1}}}\\\nn
    &\le& 4\betat(\delta)\sqrt{c_1T\Gamma_{k}(q_{r-1})} +(u_T(\delta)+1)\betat(\delta)\sqrt{\frac{c_1\Gamma_k(q_{r-1})}{q_{r-1}}}.
\end{eqnarray}

The second inequality is obtained by the choice of $t_r$. 

\textcolor{blue}{\emph{When $\Xc$ is continuous (The setting of Theorem~\ref{the:regretcont}):}}

Replacing the bound on instantaneous regret from~\eqref{eq:instcont}, for $r\ge 2$,

\begin{eqnarray}\nn
\Rc_r &\le& t_r\left(\frac{8}{q_{r-1}} + 4\betat'_{r-1}(\delta)\left(\sqrt{\frac{c_1\Gamma_k(q_{r-1})}{q_{r-1}}}+\frac{2}{\sqrt{q_{r-1}}}\right)\right)\\\nn
&\le& \lceil\sqrt{Tq_{r-1}}+u_{T}(\delta)\rceil \left(\frac{8}{q_{r-1}} + 4\betat'_{r-1}(\delta)\left(\sqrt{\frac{c_1\Gamma_k(q_{r-1})}{q_{r-1}}}+\frac{2}{\sqrt{q_{r-1}}}\right)\right)\\\nn
&\le&
4\betat'_{r-1}(\delta)\sqrt{c_1T\Gamma_k(q_{r-1})} + 8\sqrt{\frac{{T}}{q_{r-1}}} + 2\sqrt{T}\\\nn
&&~~+ (u_T(\delta)+1) \left(\frac{8}{q_{r-1}} + 4\betat'_{r-1}(\delta)\left(\sqrt{\frac{c_1\Gamma_k(q_{r-1})}{q_{r-1}}}+\frac{2}{\sqrt{q_{r-1}}}\right)\right).
\end{eqnarray}

We also note that $f(x) =\langle f, k(\cdot,x) \rangle$ (reproducing property). Thus $|f(x)|\le \|f\|_{\Hc_k}\|k(\cdot,x)\| $; i,e, $|f(x)|\le C_k k_{\max}$. Thus, the regret in the first round can be simply bounded using its length, 
\begin{eqnarray}\nn
    \Rc_1 &\le& t_1  C_k k_{\max}\\\nn
    &\le& \lceil\sqrt{T}+u_{T}(\delta)\rceil C_k k_{\max}.
\end{eqnarray}

Recall $u_T(\delta)=\Oc(\E[\tau]+\log(\frac{T}{\delta}))$ and note that, for $r>1$, 
\begin{eqnarray}
    \frac{8}{q_{r-1}} + 4\betat'_{r-1}(\delta)\left(\sqrt{\frac{c_1\Gamma_k(q_{r-1})}{q_{r-1}}}+\frac{2}{\sqrt{q_{r-1}}}\right) = \Oc(1),
\end{eqnarray}
since $q_{r-1}\ge \sqrt{T}$ for $r>1$. 

Adding up the regret over all rounds $r=1,2, \dots, R$ with $R=\log\log(T)$, we obtain 

\begin{equation}
    \Rc(T) = \Oc\left(\betat(\delta)\log\log(T)\sqrt{T\Gamma_k(T)} + C_k\E[\tau]\right),
\end{equation}

and, 
\begin{equation}
    \Rc(T) = \Oc\left(\betat'_R(\delta)\log\log(T)\sqrt{T\Gamma_k(T)} + C_k\E[\tau]\right),
\end{equation}
under the discrete and continuous $\Xc$ cases, proving Theorems~\ref{the:regret} and~\ref{the:regretcont}, respectively.

Corollary~\ref{cor:spec} immediately follows from replacing the value of $\Gamma_k$, in particular, $\Gamma_k(T)=\Oc(T^{\frac{d}{2\nu+d}}(\log(T))^{\frac{2\nu}{2\nu+d}})$, $\Gamma_k(T) = \Oc((\log(T))^{d+1})$ and $\Gamma_k(T)=\Oc(d\log(T))$, in the cases of Mat{\'e}rn, SE and linear kernels respectively ~\citep{srinivas2010gaussian, vakili2020information}. In the case of linear kernel, we emphasize the $\Oc(\sqrt{d})$ factor in $\betat'_R(\delta)$.

\end{document}